\documentclass[letterpaper, 10 pt, conference]{ieeeconf}  

\IEEEoverridecommandlockouts                              

\usepackage{graphics} 
\usepackage{epsfig} 
\usepackage{amsmath} 
\usepackage{amssymb}  
\usepackage[caption = false]{subfig}
\usepackage{hyperref}
\usepackage{caption}

\title{\LARGE \bf
Face Images as Jigsaw Puzzles: Compositional Perception of\\ 
Human Faces for Machines Using Generative Adversarial Networks
}

\author{Mahla Abdolahnejad$^{1}$ and Peter Xiaoping Liu$^{2}$
\thanks{The authors are with the Department of Systems and Computer Engineering,
Carleton University, Ottawa, ON K1S 5B6, Canada.}
\thanks{$^{1}$
        {\tt\small mahlaabdolahnejadbah@cmail.carleton.ca}}%
\thanks{$^{2}$
        {\tt\small xpliu@sce.carleton.ca}}%
}

\begin{document}

\maketitle
\thispagestyle{empty}
\pagestyle{empty}

\begin{abstract}

An important goal in human-robot-interaction (HRI) is for machines to achieve a close to human level of face perception. One of the important differences between machine learning and human intelligence is the lack of compositionality. This paper introduces a new scheme to enable generative adversarial networks to learn the distribution of face images composed of smaller parts. This results in a more flexible machine face perception and easier generalization to outside training examples. We demonstrate that this model is able to produce realistic high-quality face images by generating and piecing together the parts. Additionally, we demonstrate that this model learns the relations between the facial parts and their distributions. Therefore, the specific facial parts are interchangeable between generated face images.

\end{abstract}

\section{INTRODUCTION}

Face perception is essential for social interactions. Therefore a significant topic in human-robot-interaction (HRI) is to close the gap between humans' and robots' face perception ability. Several studies have been focused on improving different aspects of face perception for robots. For instance, recognizing human emotions and expressions \cite{siqueira2018ensemble}\cite{cosentino2018group}\cite{saxena2019deep}, gaze following \cite{zhang2016optimal}\cite{saran2018human}, and engagement estimation from face images \cite{rudovic2018culturenet}\cite{kompatsiari2019measuring}. One important subject in cognitive science and artificial intelligence (AI) is compositionality. This is the idea that a new concept can be constructed by combining the primitive components. Nevertheless, the lack of compositionality is one of the important differences between machine learning and human intelligence \cite{abdolahnejad2020deep}\cite{lake2017building}. 

\begin{figure}[t]
\begin{center}
\subfloat{\includegraphics[trim= 0 310 0 0, clip, scale=0.25]{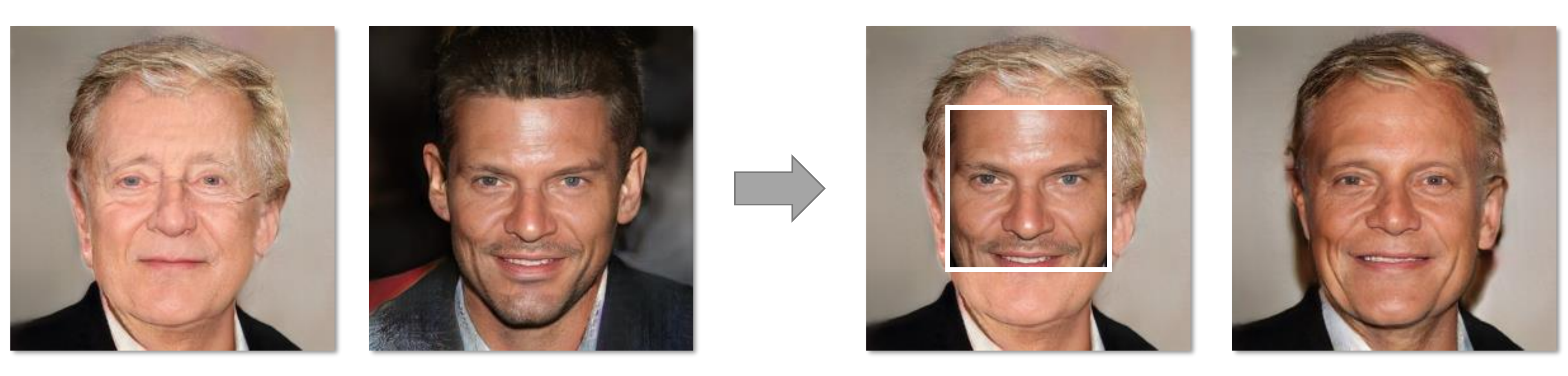}}
\\
\subfloat{\includegraphics[trim= 0 250 0 0, clip, width=3.3in]{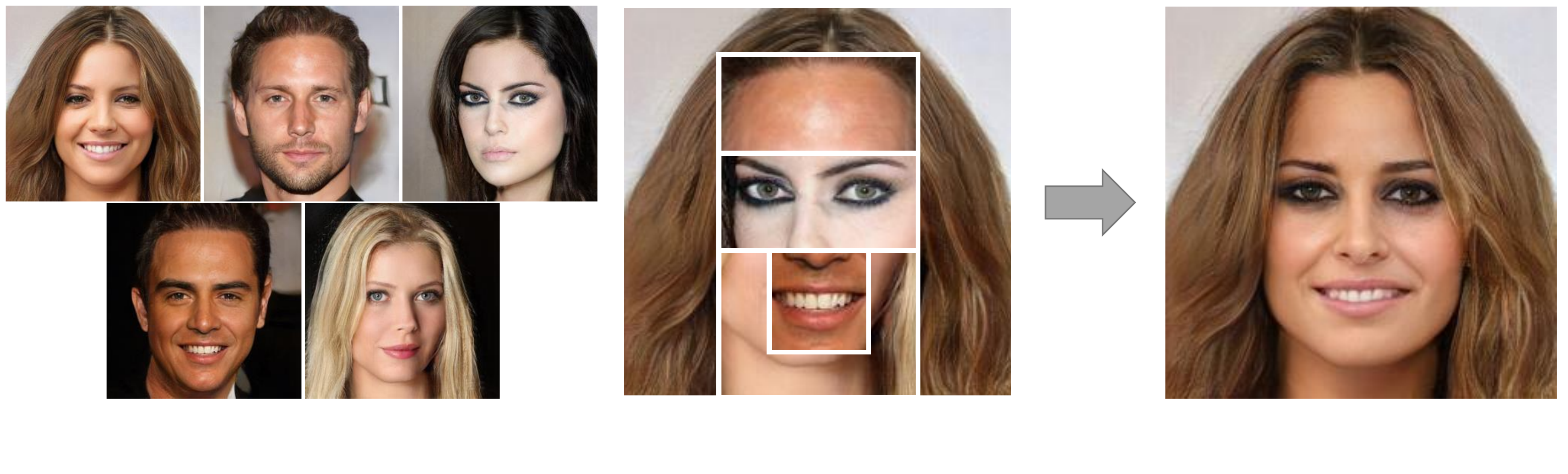}}
	\caption{The compositional GAN face perception model is able to adapt an entire puzzle in order to fit in the pieces from different puzzles.}
\label{puzzle}
\end{center}
\end{figure}

Visual concepts displayed in images can be represented as compositions of parts and relations. This provides a base for constructing new visual concepts. For example, face images are composed of components such as eyes, eyebrows, nose, and mouth placed at specific positions within the face structure. Accordingly, a face image can be divided into parts with fixed positions and sizes each containing specific facial components. Our goal is to improve the ability of machines to understand faces by representing face images as compositions of such parts. For example, imagine having thousands of jigsaw puzzles of different face images. Assuming that all of the jigsaw puzzles have the same number of pieces and the same arrangement of the pieces, our model learns the distribution for each puzzle piece. Additionally, by replacing one or more pieces of a jigsaw puzzle with pieces from the same location of a different puzzle, the model allows the entire puzzle to adapt itself in order to fit in the piece from the different puzzle (see Figure \ref{puzzle}). By defining each puzzle piece to only contain a specific facial component, learning a separate distribution for each piece leads to the disentanglement of facial components in the representation space. Such implementation of compositionality is close to the human way of understanding and describing visual concepts \cite{hoffman1983parts} and makes it possible to construct an infinite number of whole representations from a finite set of parts representations.

To summarize, the main contribution of this paper is to enable machines to learn the distribution of face images as a composition of distributions of facial parts. We present a straightforward method to improve a standard generative adversarial networks (GANs) architecture such that the modified architecture is able to learn multiple localized representations for a face. We demonstrate that our compositional GAN model not only learns the representations for facial parts but also the relations between them. Therefore it can generate realistic whole images given any combination of samples from parts distributions.

\section{Related Work}

There are variations of GANs that implement some form of compositionality. For instance, Spatial Transformer GANs (ST-GANs) \cite{lin2018st} utilizes Spatial Transformer Networks (STNs) as the generator in order to produce natural images by combining a foreground object with a background image. They propose an iterative STN warping scheme and a sequential training strategy in order to visualize how accessories like eyeglasses would look when matched with real face images. Bau et al. \cite{bau2018gan} use a segmentation-based network dissection method in order to identify a group of activation units that are closely related to object concepts. By zeroing these activations in the output image it is possible to remove that object from the image. Also it is possible to insert these object concepts into output images. Collins et al. \cite{collins2020editing} make local, semantically-aware edits to GANs generated human faces by introducing a novel way for manipulation of style vectors of StyleGAN \cite{karras2019style}. They demonstrate that performing clustering on hidden layers activations results in the disentanglement of semantic objects. Azadi et al. \cite{azadi2020compositional} propose a Composition-by-Decomposition network to compose a pair of objects. Given a dataset of objects $A$ images, a dataset of object $B$ images, and a dataset of both objects $A$ and $B$ images (all datasets labeled with masks), the model can receive a sample from $A$ and a sample from $B$ and generate a realistic composite image from their joint distribution. In comparison, our method does not use spatial transformations, or manipulation and dissection of the hidden layer activations. It also does not require masks for different objects within the image. However, it introduces a type of structure and hierarchy in order to improve the training of GANs and to achieve the disentanglement of concepts/objects in the latent variable space.
 
There are a large group of methods that add structure and hierarchy to the training of GANs. For instance, LAPGAN \cite{denton2015deep} decomposes the image into a set of band-pass images with one octave space between them with a separate GANs model at each level of the pyramid for generating images in a coarse-to-fine manner. SGAN \cite{huang2017stacked} proposes a top-down stack of GANs in which the GANs model at each level is conditioned on a higher-level representation responsible for generating a lower-level representation. Progressive Growing of GANs (PGAN) \cite{karras2017progressive} begins the training with smaller networks for the generator and discriminator and then grows their size progressively by adding layers during training. The hierarchy and structure proposed in all of the methods mentioned are coarse-to-fine and are based on first building the whole image coarse structure and then adding more details to it. Contrary to these methods, we attempt to add a compositional hierarchy to the training of GANs. More precisely, we would like to divide an image into smaller parts and learn distinct representations for each part. As a result, the model will be able to produce whole images by sampling from the learned distributions for parts and then generating the parts and piecing them together.

\begin{figure*}[h]
\begin{center}
\subfloat{\includegraphics[trim=0 90 380 0, clip, scale=0.32]{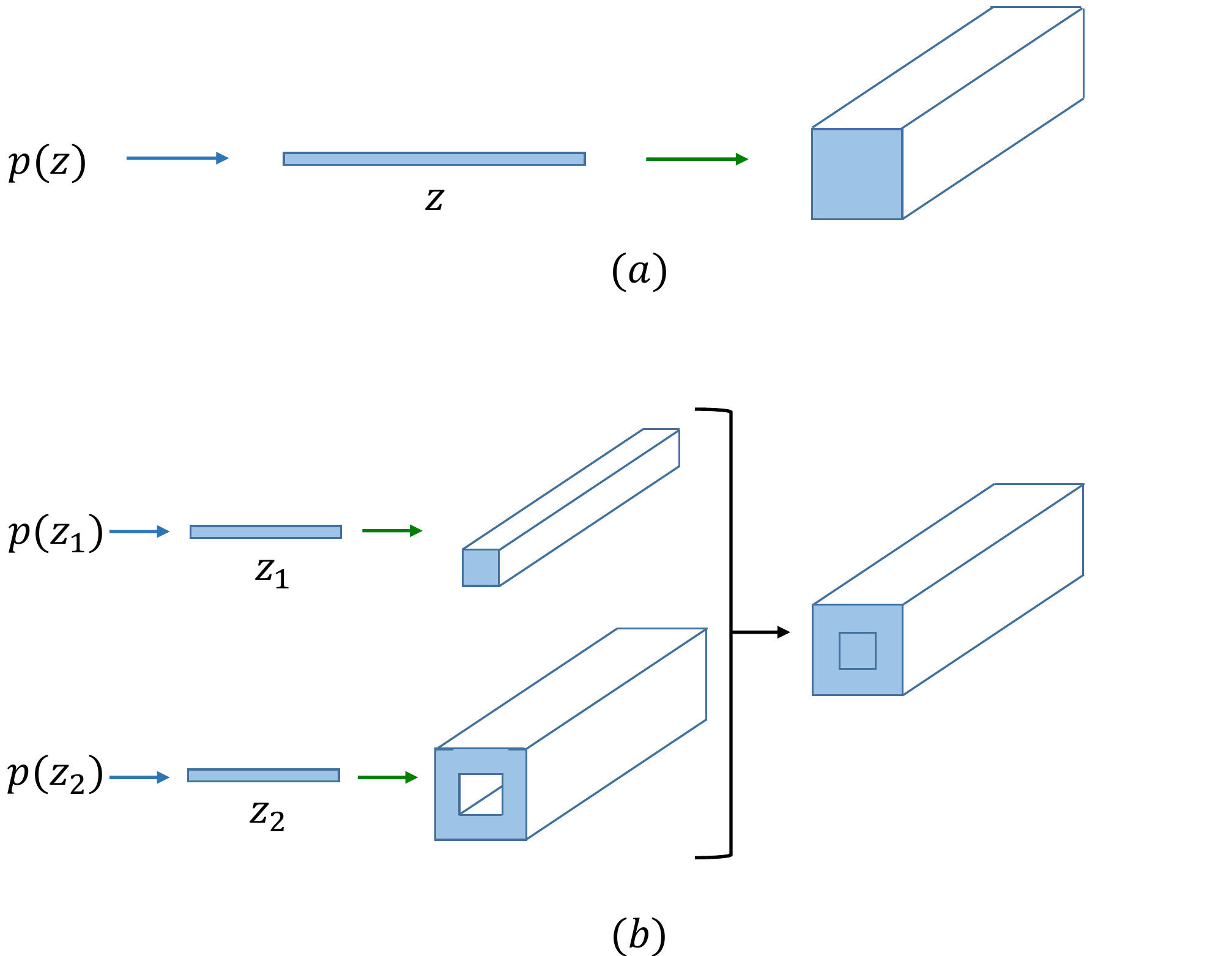}}
\subfloat{\includegraphics[trim=0 90 180 0, clip, scale=0.32]{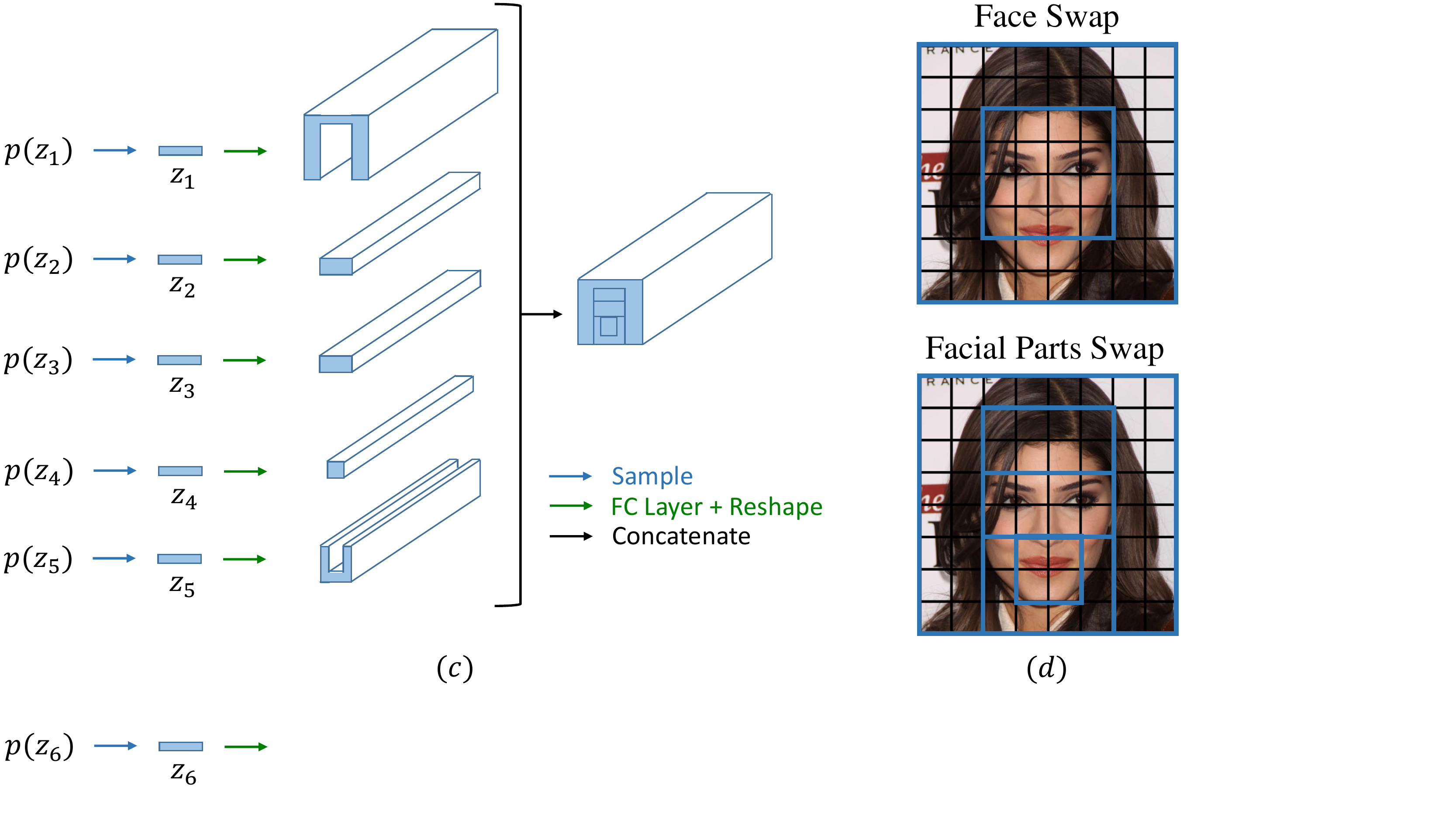}}
	\caption{(a) The first layer of generator network in a standard GANs architecture. (b) The first layer of generator network in order for the compositional GANs model to generate face images as composition of two parts; one for face and one for everything else. (c) The first layer of generator network in order for the compositional GANs model to generate face images as composition of five distinct facial parts. (d) The illustration of how parts are defined (top) for face-swap task, and (bottom) for a more elaborate division of face images into five facial parts.}
\label{firstlayer}
\end{center}
\end{figure*}

\section{Generating Images Similar to Building Jigsaw Puzzles}

\subsection{Background: Standard GANs}

GANs \cite{goodfellow2014generative} is a remarkable method for modeling data distributions. Instead of explicitly modeling and solving the dataset density function, GANs focus on generating samples from this distribution. GANs’ main idea is to first sample from a known latent distribution and then transform this sample into a sample of training distribution. Two networks are involved in a GANs model; discriminator network and generator network.  Given a data distribution $x \sim p_{data}(x)$, $x \in \mathcal{X}$, the generator network maps samples of a latent prior $z \sim p_z(z)$ to samples of the training distribution $G: \mathcal{Z} \rightarrow \mathcal{X}$. The generator network $G(z;\theta_g)$ implicitly defines a distribution over data; $p_g(x)$. The discriminator network $D(x; \theta_d)$ receives the real data samples and also the samples synthesised by generator network and estimates the probability of whether they are real or fake via a score $D: \mathcal{X} \rightarrow {\rm I\!R}$. The training process of GANs constantly alternates between training of the discriminator network and training of the generator network. This can be interpreted as the two networks playing a minimax game with the value function

\begin{align}
\underset{G}{\mathrm{min}} \, \underset{D}{\mathrm{max}} \: V(D,G) &= {\mathbb{E}}_{x \sim {p_{data}}(x)} [\mathrm{\log} \: D(x; \theta_d)] \\
&+ {\mathbb{E}}_{z \sim p_z(z)} [\mathrm{\log} (1 - D(G(z; \theta_g); \theta_d))]. \nonumber
\end{align}

This minimax game will eventually converge to an equilibrium state in which the samples generated by generator network are identical to real training data and discriminator network assigns probability of $0.5$ to every input regardless of whether it is real or fake.

\subsection{Our Compositional GANs}

To achieve a compositional GANs, we add an assumption to the standard GANs model that the latent distribution $p_z(z)$ is composed of $K$ distinct distributions $\{p_{z_i}(z_i)\}_{i=1}^{K}$ and a syntactic method $R_z$. Assuming that there is a function $\mu$ that pairs up each $z_i$ with an image part $x_i$ and also $R_z$ (i.e. the way $z_i$s are combined) with $R_x$ (i.e. the way $x_i$s are combined) 
\begin{align}
x_1 = \mu(z_1), ..., x_k = \mu(z_k), R_x = \mu(R_z),
\end{align} 

\noindent then the system is compositional if and only if there is a function $f()$ such that every (syntactically) complex item $x = \mu(z)$ in the syntactic system is a function of syntactic parts $x_i = \mu(z_i)$ and the way they are combined $R_x=\mu(R_z)$ \cite{pelletier2017compositionality}.

The generator of our compositional GANs model performs the following mapping
\begin{align}
G: \{\mathcal{Z}_i\}_{i=1}^{K}, R_z \xrightarrow{\mu} \{\mathcal{X}_i\}_{i=1}^{K}, R_x \xrightarrow{f} \mathcal{X}.
\end{align}

\noindent It means that the generator network learns the mapping $\mu()$ from the samples of parts priors $z_i$ and the relationship among them $R_z$ to the image parts $x_i$ and the relationship among them $R_x$. It also learns the mapping $f()$ from image parts $x_i$ and the relationship among them $R_x$ to whole image $x$.

\begin{align}
G(z, \theta_g) = f(\mu(z_1, ..., z_K, R_z; \theta_\mu); \theta_f) = p_g(x)
\end{align}

This can be interpreted as a group of generators working in parallel to generate the image parts. However, since the adversarial loss is defined for the realistic whole images and not the individual image parts, the model also learns the relations between the parts in order to make the whole pieced-together image comparable with real samples from the dataset.

\section{Experiment}

In this section, we detail the building and training of compositional GANs for the purpose of learning the distribution of face images. Further experiments and results including compositional GANs implemented for handwritten digits and bedroom images along with our code can be found here: \href{https://github.com/MahlaAb/puzzlegan}{https://github.com/MahlaAb/puzzlegan}.

\subsection{Implementing Parts and Compositions}

Typically in GANs architecture the first layer of the generator network is for generating a volume $w \times h \times c$ by passing the input latent vector $z$ through a fully-connected layer followed by a reshaping operation (Figure \ref{firstlayer}(a)). The purpose of this transition from a $1$-D vector to a $3$-D volume is to configure the layers input/output shapes compatible with the subsequent convolutional layers. Our modifications to the standard GANs model focuses on this component of the generator network. We replace the single input vector $z$ with multiple input vectors $\{z_i\}_{i=1}^{K}$ with each passing through a distinct fully-connected layer followed by a distinct reshaping operation. The results are multiple volumes $\{v_i\}_{i=1}^{K}$ which can be put together to create $v = w \times h \times c$ which is the input to the first convolutional layer. Figure \ref{firstlayer}(b) and \ref{firstlayer}(c) respectively illustrate the way these modifications are performed for the purpose of a face swap task and a facial parts swap task.

\begin{figure*}[h]
\begin{center}
\hspace{10pt} \textbf{Generator}  \hspace{180pt} \textbf{Discriminator} \\
\subfloat{\includegraphics[trim=0 80 125 92, clip, scale=0.21]{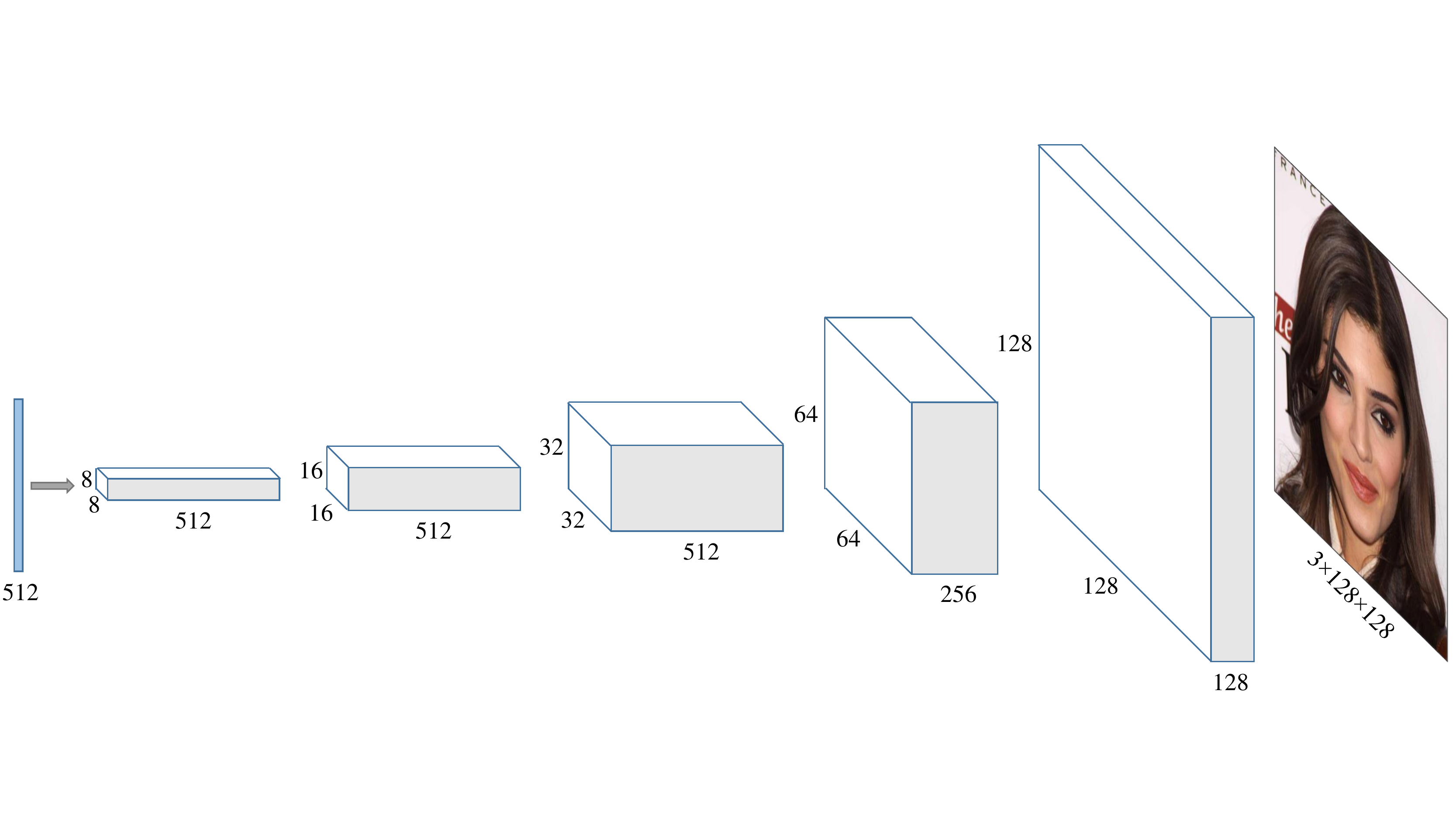}}
\subfloat{\includegraphics[trim=660 80 80 92, clip, scale=0.21]{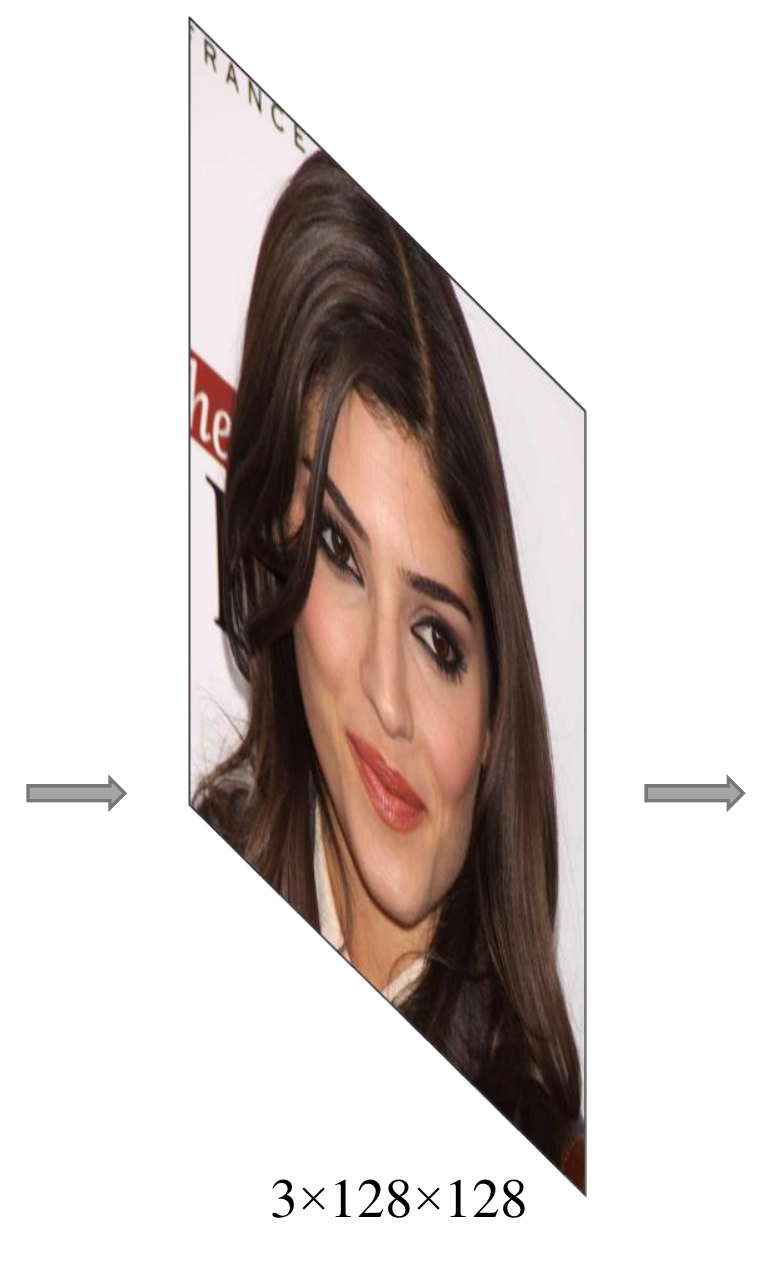}}
\subfloat{\includegraphics[trim=125 80 0 92, clip, scale=0.21]{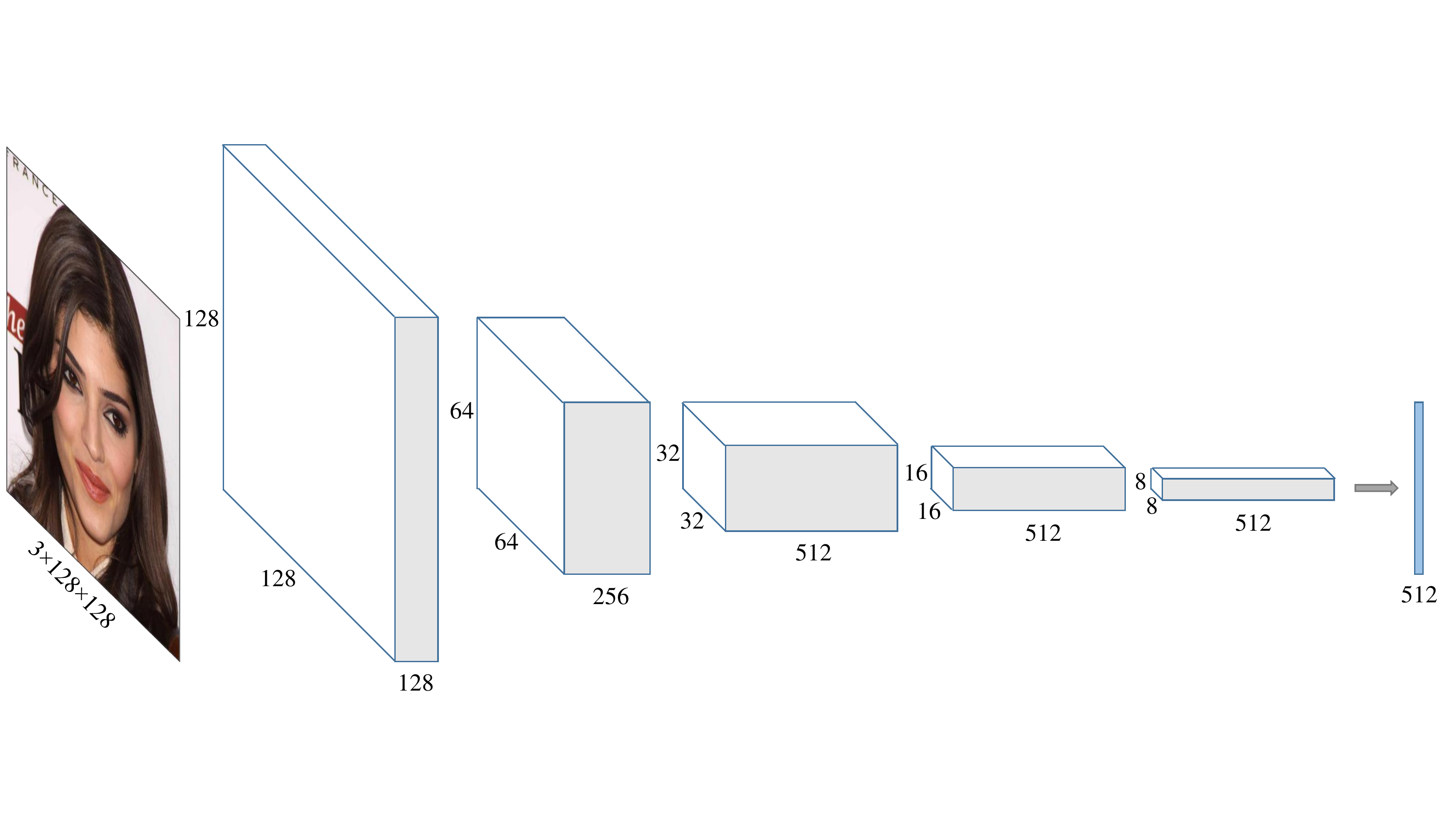}}\\
\subfloat{\includegraphics[trim=0 80 125 92, clip, scale=0.21]{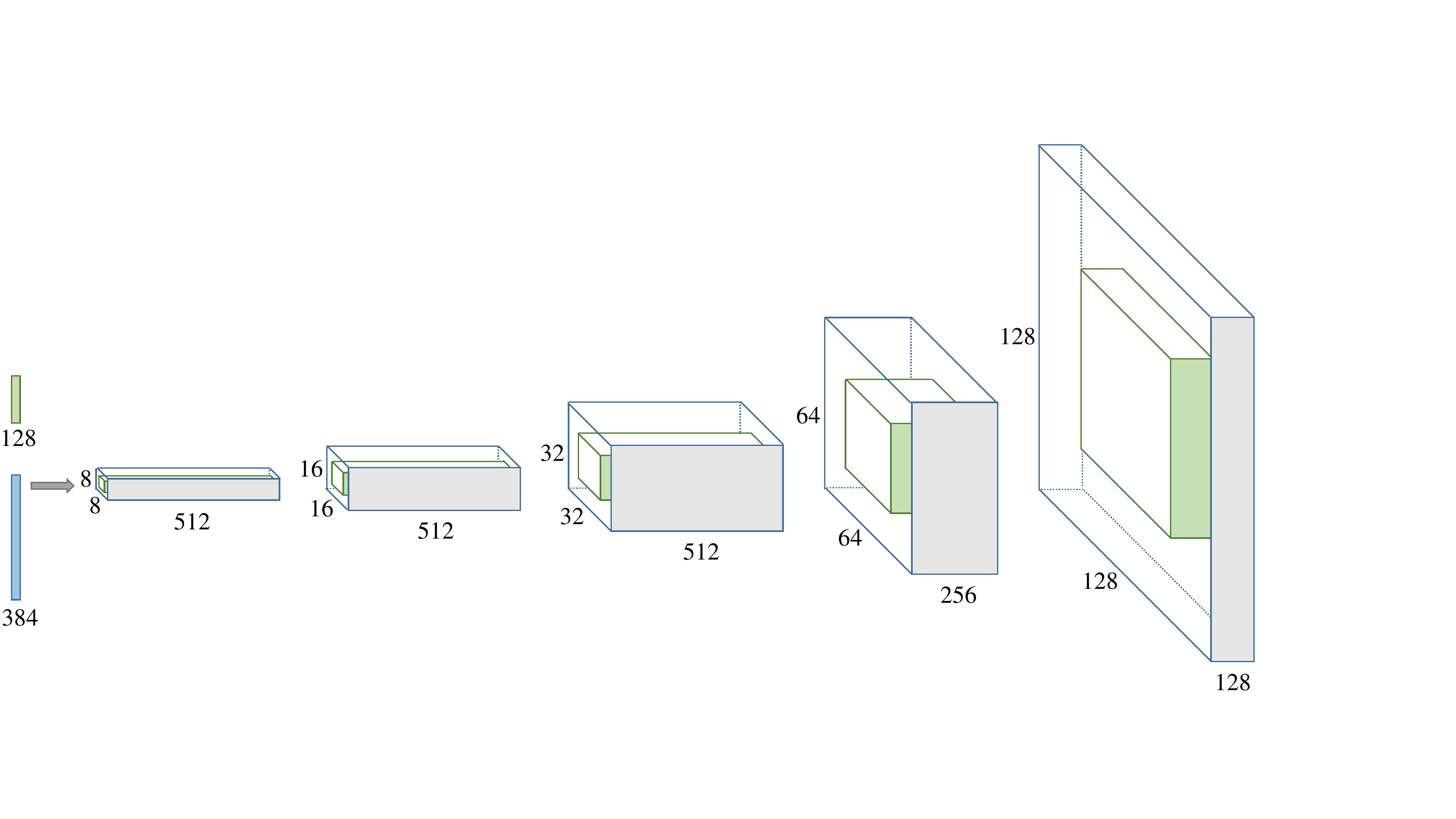}}
\subfloat{\includegraphics[trim=660 80 80 92, clip, scale=0.21]{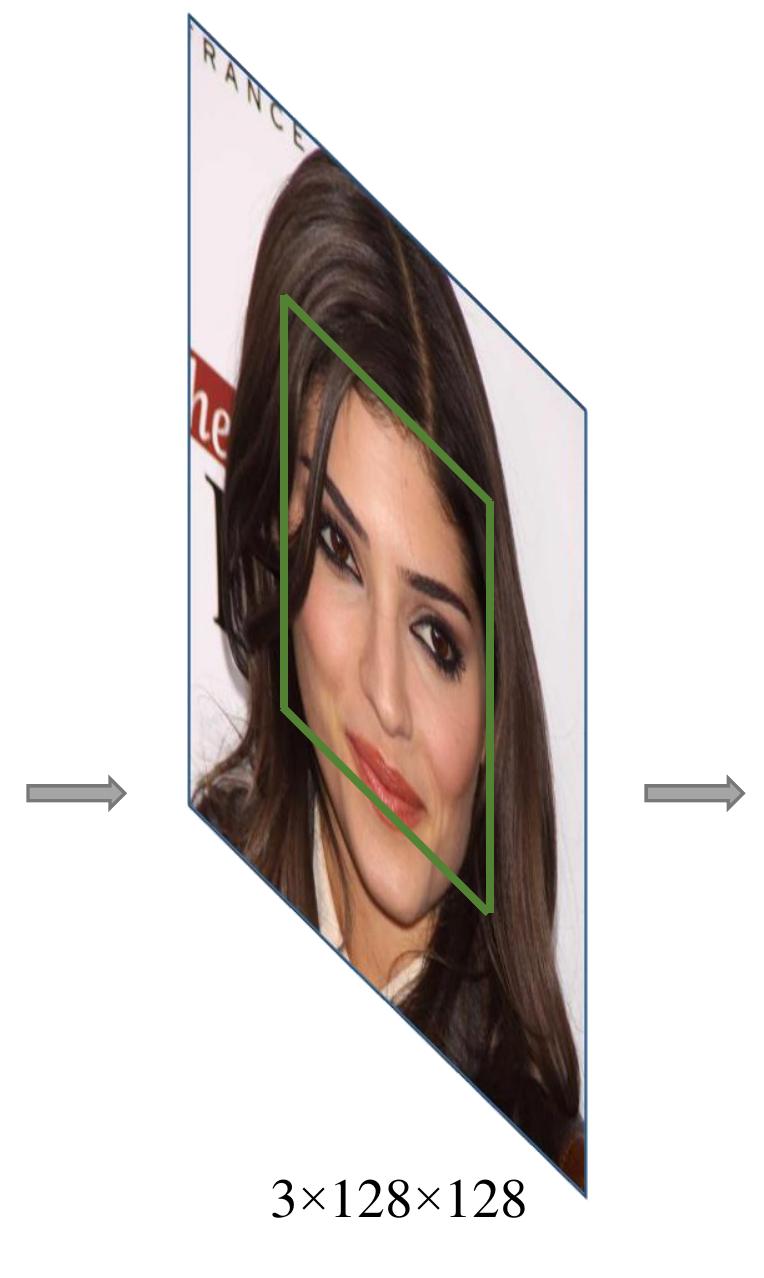}}
\subfloat{\includegraphics[trim=125 80 0 92, clip, scale=0.21]{D}}\\
\subfloat{\includegraphics[trim=0 60 125 92, clip, scale=0.21]{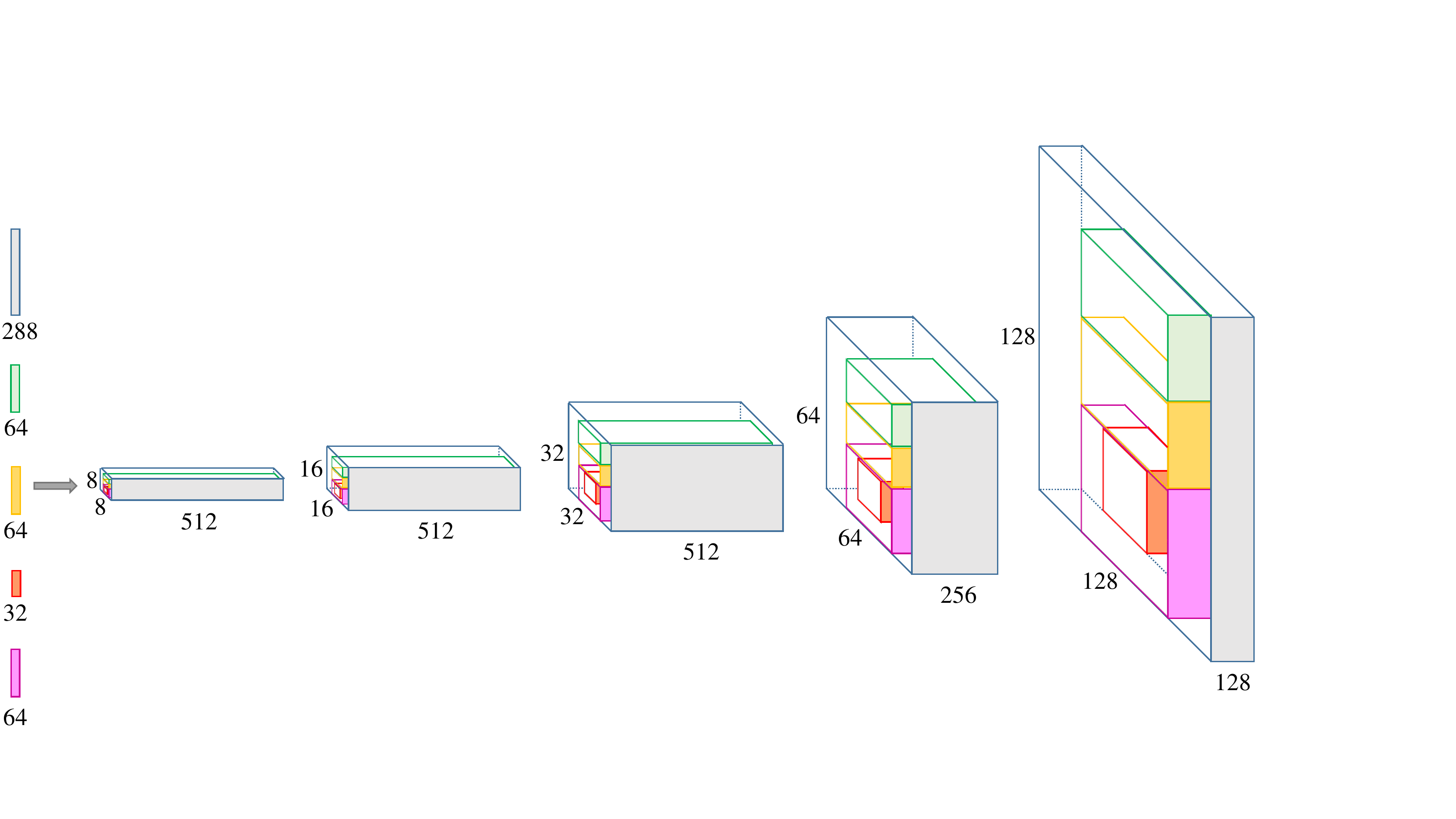}}
\subfloat{\includegraphics[trim=660 60 80 92, clip, scale=0.21]{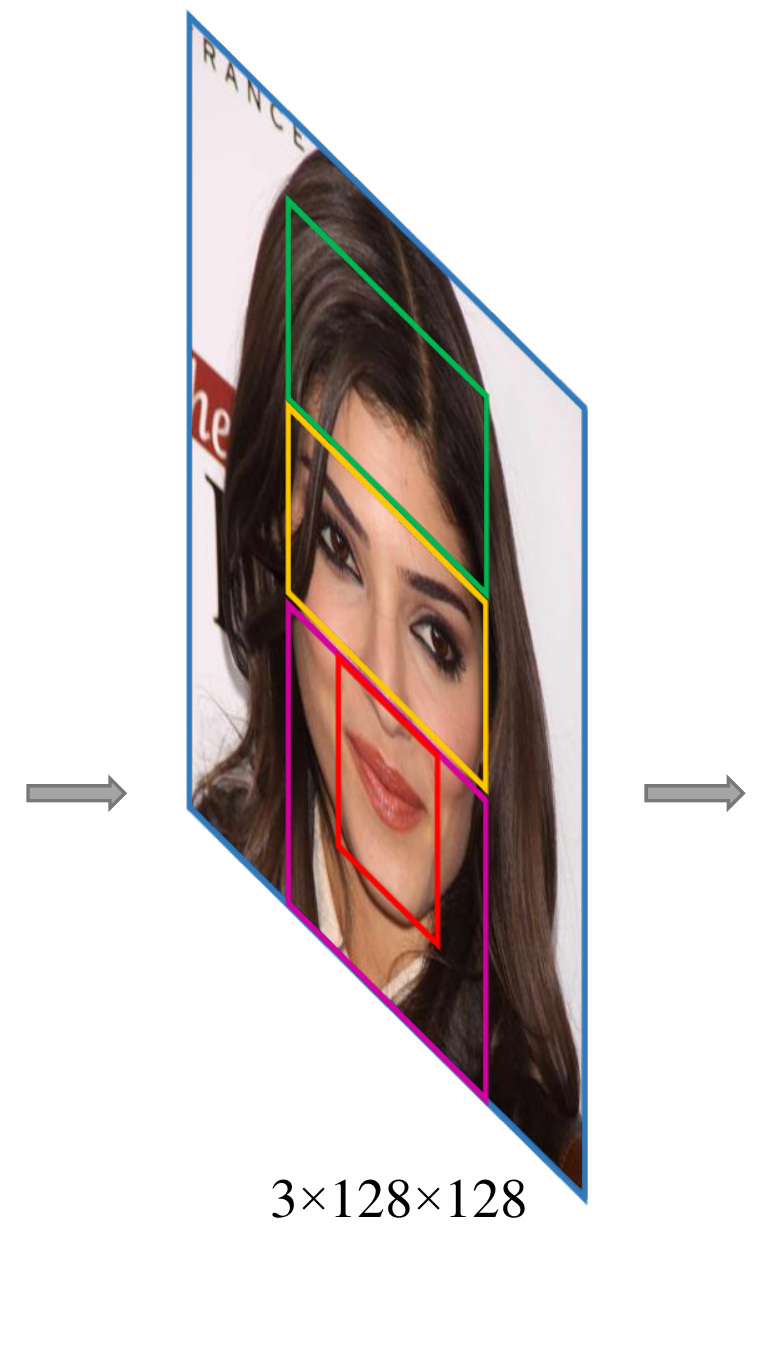}}
\subfloat{\includegraphics[trim=125 60 0 92, clip, scale=0.21]{D}}\\
	\caption{The schematic view of (top) the generator network and the discriminator network of a standard GANs model, (middle) the generator network and the discriminator network of GANs modified in order to have two distinct parts; one for the face and one for everything else, (bottom) the generator network and the discriminator network of GANs modified in order to have five distinct parts each containing specific facial components.}
\label{architect}
\end{center}
\end{figure*}

Convolutional layers operate by scanning an input volume row by row starting from the top left corner. Therefore, the arrangement of the volumes $v_i$ is carried from the input to the output of a convolutional operation. Similarly, as it is illustrated in Figure \ref{architect}, the arrangement of the volumes $v_i$ from the first layer carries throughout the entire network until the output image is generated without requiring any further modifications to the architecture. In summary, the choice of shapes and the manner of arranging the volumes $v_i$ into the larger volume $v$ in the first layer of the generator network determines which part of the output images each latent variable $z_i$ is responsible to represent. Returning to our jigsaw puzzle analogy, it can be stated that in order to generate an image similar to building a jigsaw puzzle we need to build the input of the first convolutional layer similar to building a Lego block with bricks of different sizes and shapes.

\subsection{Dividing Faces into Parts}

When a human is asked to describe a face image they have the ability to divide it into parts and then describe each part. We would like our model to have a similar way of dividing face images into smaller parts. Clearly, for this to be achieved an alignment among the training samples is required. However, this is not a very strict notion of alignment. More precisely, the model requires the specific concepts to be located in approximately the same position within all training examples. Therefore, only by dividing training examples into predefined parts and learning a representation for each part we can assure that the representation learned for a part describes a single concept and the model succeeds in learning it.

Here we first divide a face image using an $8 \times 8$ grid and then define the parts as shown in Figure \ref{firstlayer}(d). The top figure illustrates how parts are defined for a face-swap task meaning that the face image is divided into two parts only; one for the face and one for everything else. The bottom figure shows a more elaborate division of face images into multiple facial parts. As it can be seen from this figure, five distinct parts are defined to contain 1) the background and hair, 2) hairline and forehead, 3) eyes, 4) nose and mouth, and 5) overall shape of the face. The dataset used for training our model is CelebA-HQ dataset \cite{karras2017progressive}. It contains 30,000 HD aligned photographs of celebrity faces. Additionally, we used the Progressive Growing of GANs (PGAN) \cite{karras2017progressive} as a base for implementing our model.

\section{Qualitative evaluation}



\begin{figure*}[h!]
\begin{center}
\setlength{\unitlength}{1in}
\subfloat{\includegraphics[width=0.34in]{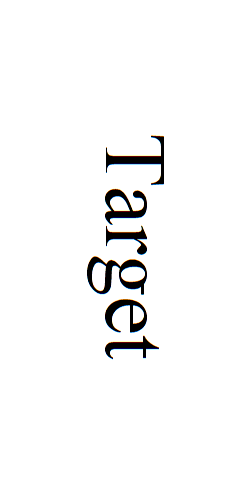}}
\subfloat{\includegraphics[width=0.7in]{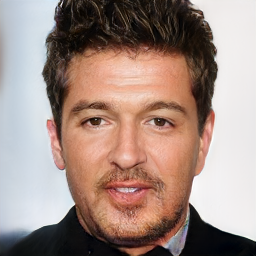}}
\hspace{6pt}
\subfloat{\includegraphics[width=1.3in]{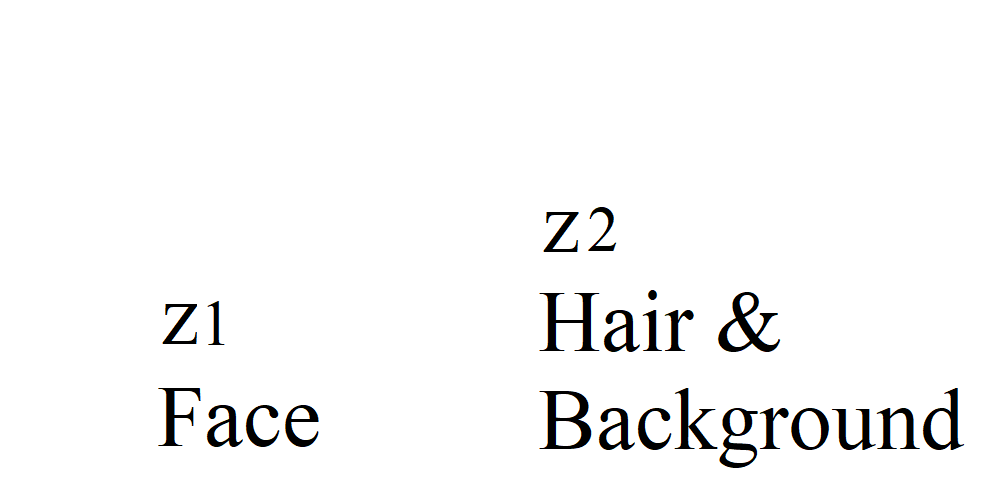}}
\hspace{13pt}
\subfloat{\includegraphics[width=0.7in]{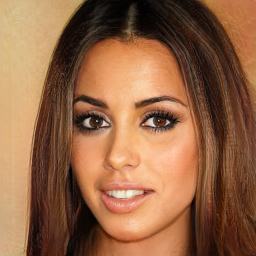}}
\hspace{5pt}
\subfloat{\includegraphics[width=3.5in]{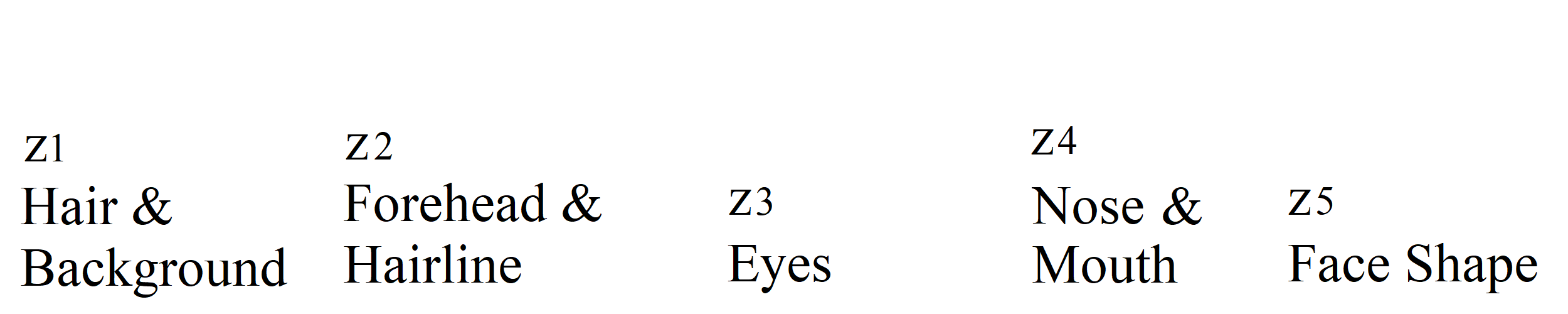}}
\\
\subfloat{\includegraphics[width=0.34in]{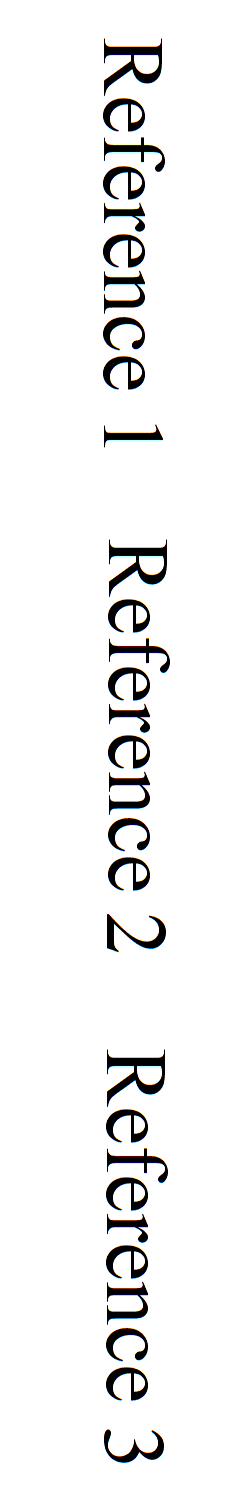}}
\subfloat{\includegraphics[width=2.2in]{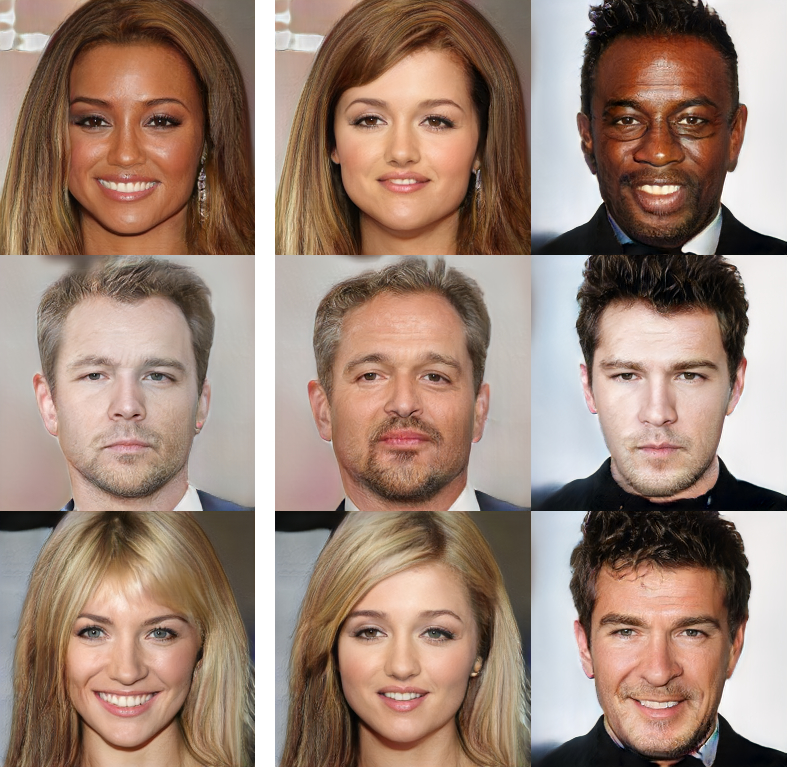}}
\hspace{8pt}
\subfloat{\includegraphics[width=4.3in]{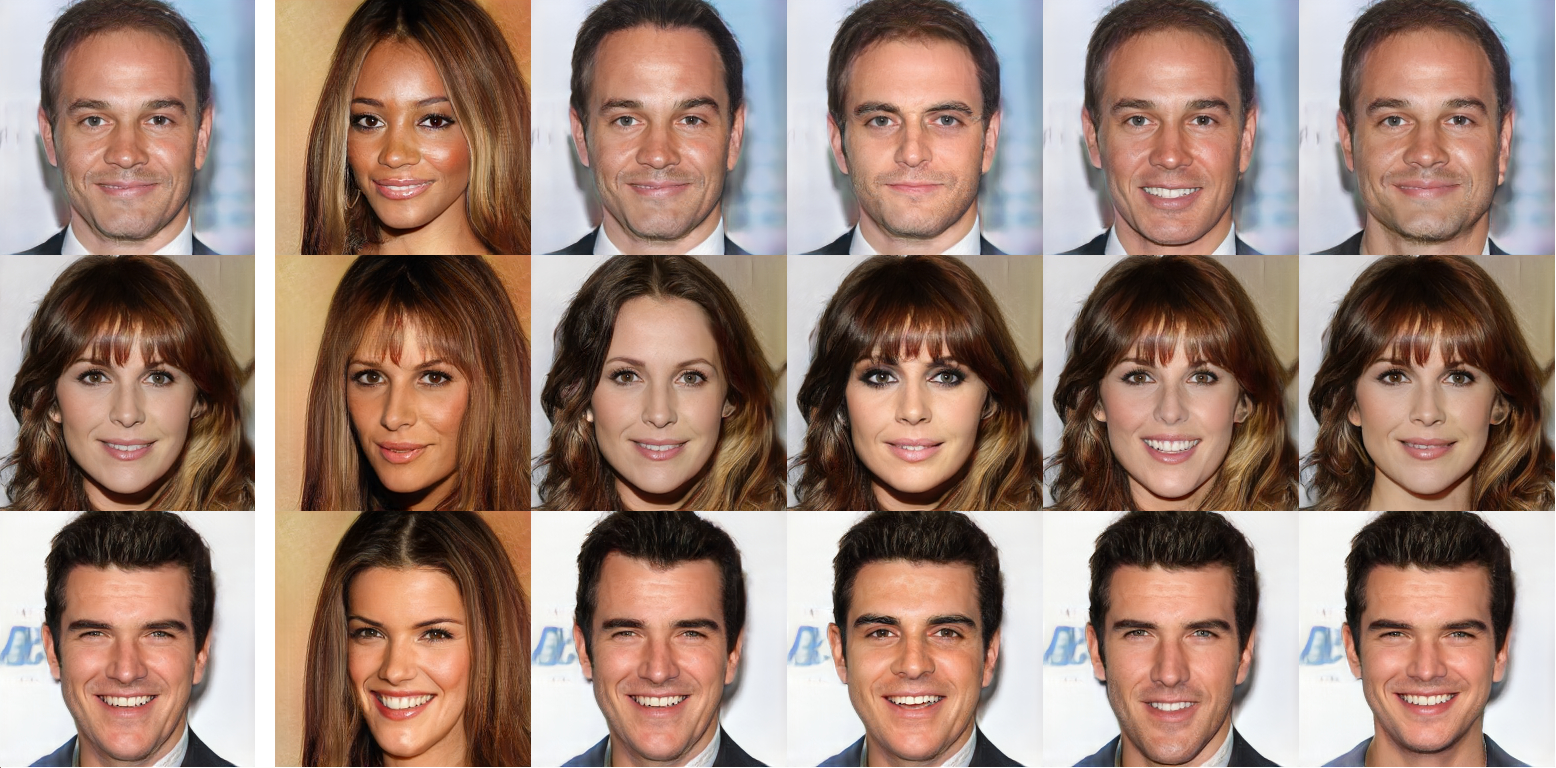}}
	\caption{Example generated faces given the combination of priors from a target image and a reference image (left) by the face swap compositional GAN model, and (right) by the facial parts compositional GAN model.}
\label{fp_mixings}
\end{center}
\end{figure*}

A compositional GANs model makes it possible to control the parts in the generated output faces. Figure \ref{fp_mixings} displays examples of the faces generated by our model and also the results of exchanging a single prior sample $z_i$ between two face images. The faces in the top row of this figure are the example generated faces that are used as target images, while the remaining three images below each target face show the reference images that are used as the source of all priors but one. The remaining faces are the generated faces by the compositional GAN model given the combination of priors from a target image and a reference image. Another example is Figure \ref{puzzle}(bottom) in which five prior samples from five different source faces are combined to generate one output face.

These images demonstrate that the model was able to generate realistic face images and more importantly, it learned the relations among the parts. For example if a $z_2$ sample representing blonde hairline is combined with a $z_1$ sample representing black hair, the model is able to match the colors. Another example is matching the eyebrows color in $z_3$ with the hair color in $z_1$. We also observed that the $z_i$s that are responsible for smaller areas of the final image such as $z_4$ cause very subtle changes in the output image while $z_i$s that are accounting for larger areas of the image such as $z_1$ create more dominant influence. For example the hairstyle in $z_1$ dictates the gender in the output image.

\section{Quantitative analysis}

\subsection{Photorealism}

An important criteria for evaluating a generative model is the measure of photorealism in the synthesised images. Frechet inception distances (FID) is a measure of how close the generated images are to the real samples and is computed by feeding a set of real images and a set of synthesised images to the Inception-V3 network and computing the difference between inception representations of the two groups. Table \ref{table1} demonstrates that removing one block from the generator and the discriminator of the PGAN model, which is required for our compositional GAN model, does not affect the quality of generated images negatively.

Additionally, Table \ref{table2} compares the quality of generated images by a standard PGAN model with an $8 \times 8$ first block against the quality of generated images by our face swap compositional GANs and facial parts compositional GANs models. The table provides evidence that the PGAN model can be modified in order to achieve a compositional GANs model without sacrificing the photorealism of the generated images.

\begin{table}[h]
\begin{center}
\begin{tabular}{l p{3cm}}
\hline
Model & FID Score\\
\hline
\hline
PGAN with a 4$\times$4 first block & 48.4197 \\
\hline
PGAN with an 8$\times$8 first block & 46.7152 \\
\hline
\end{tabular}
\caption{Comparison of FID scores (lower is better) for 50,000 generated images at $128 \times 128$  compared to CelebAHQ training samples. The networks are trained using 10 million images.}
\label{table1}
\end{center}
\end{table}

\begin{table}[h]
\begin{center}
\begin{tabular}{l p{3cm}}
\hline
Model & FID Score\\
\hline
\hline
PGAN with an 8$\times$8 first block & 9.6055 \\
\hline
Face Swap Compositional GANs & 10.4772 \\
\hline
Facial Parts Compositional GANs & 9.7592 \\
\hline
\end{tabular}
\caption{Comparison of FID scores (lower is better) for 50,000 generated images at $256 \times 256$ compared to CelebAHQ training samples. The networks are trained using 10 million images.}
\label{table2}
\end{center}
\end{table}

\subsection{Relationship Among The Parts}

\begin{figure*}[h!]
\begin{center}
\subfloat[]{\includegraphics[trim=0 0 580 40, clip, width=0.9in]{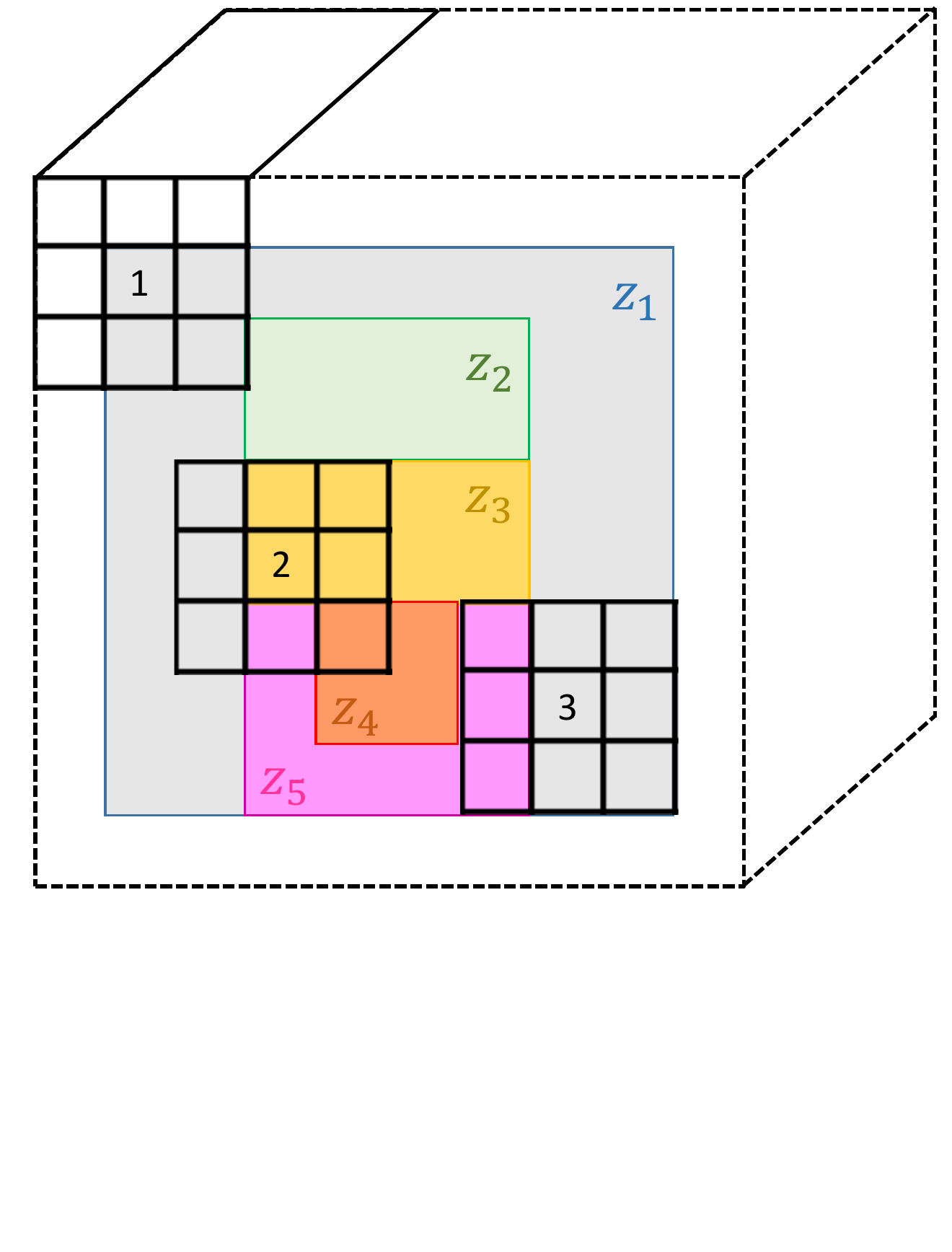}}
\hspace{40pt}
\subfloat[Face Swap]{\includegraphics[trim=0 0 650 0, clip, scale=0.26]{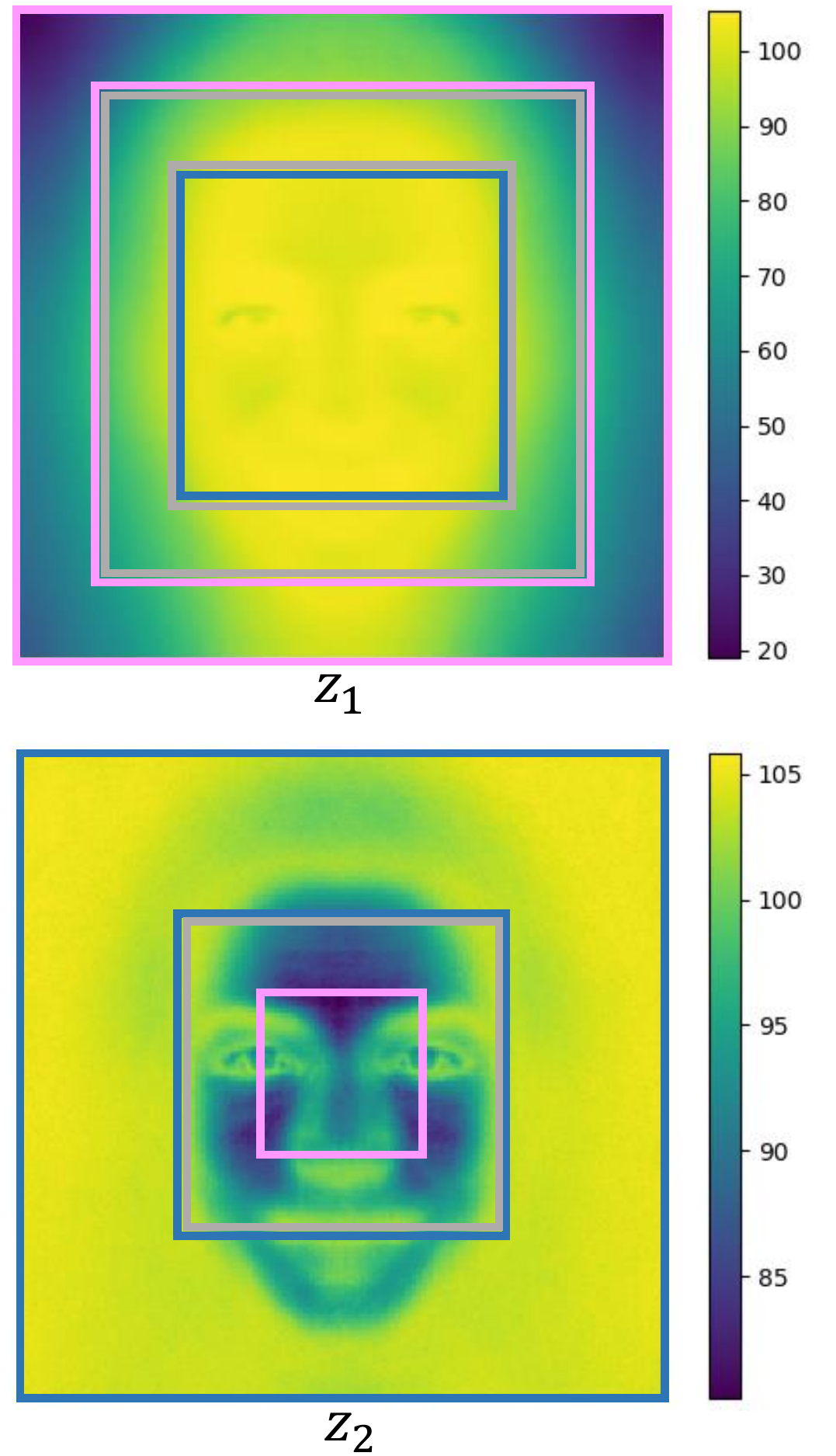}}
\hspace{10pt}
\subfloat[Facial Parts Swap]{\includegraphics[trim=0 0 0 0, clip, scale=0.26]{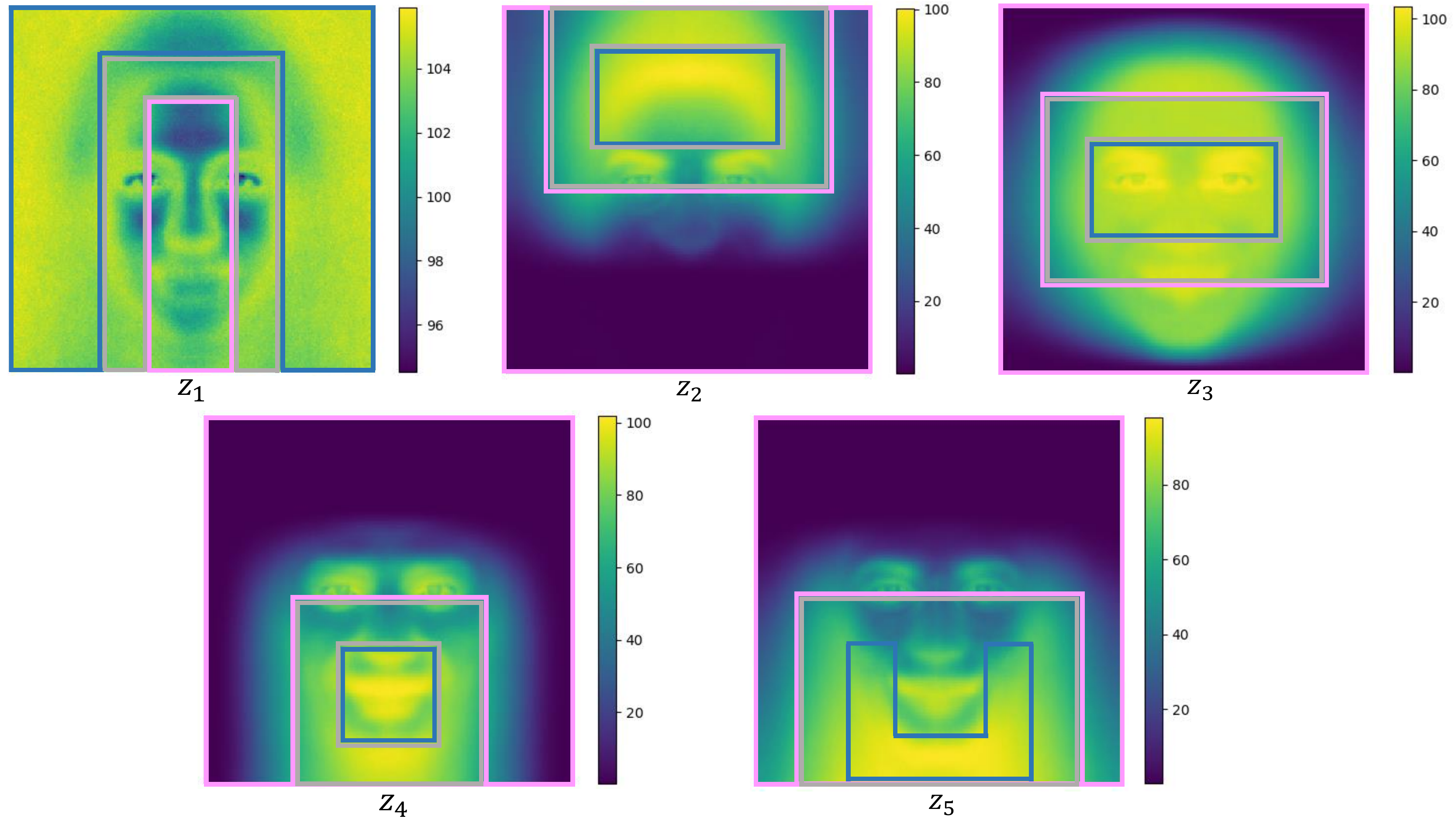}}
	\caption{(a) Illustration of a convolution operation with filter size of $3 \times 3$ and SAME padding scanning through an $8 \times 8 \times c$ block with five distinct divisions. $1$, $2$, and $3$ respectively display examples of pixels generated influenced by only one part prior, four different priors, and two different priors. (b) and (c) Mean squared-error (MSE) heatmaps computed between 50,000 generated images and their edited counterparts by replacing a single $z_i$ only. }
\label{heatmaps}
\end{center}
\end{figure*}

Figure \ref{heatmaps} (a) illustrates the first block of the generator network of our facial parts compositional GANs. The block size is $8 \times 8 \times c$ and is being scanned by a convolutional $3 \times 3$ filter. Three example positions of the filter are illustrated in the figure. As evidenced, there are regions that are only affected by one latent prior. For example the pixel labeled $1$ in the figure is generated influenced by $z_1$ only. In contrast, some regions are generated with more than one latent prior influencing them. For example, the pixel labeled $2$ is generated influenced by $z_1$, $z_3$, $z_4$, and $z_5$, and the pixel labeled $3$ is generated influenced by $z_4$ and $z_1$. These boundary regions between parts are in fact responsible for learning the interlocking of different parts.

Figure \ref{heatmaps} (b) and (c) show the Mean squared-error (MSE) heatmaps computed between 50,000 generated images and their edited counterparts by replacing a single $z_i$ only. In each heatmap, the area outlined in blue is the area that the replaced $z_i$ is responsible to represent. Meanwhile, the area outlined in grey is the interlocking area between the replaced $z_i$ and the other parts priors. Finally, the area outlined in pink is the area completely outside the influence of the replaced prior $z_i$. The heatmaps exhibit that the inside area has the largest MSE, the outside area has the lowest MSE, and with the interlocking area’s MSE in between.

\begin{figure}[h!]
\begin{center}
\subfloat{\includegraphics[trim=30 10 40 25, clip, width=0.8in]{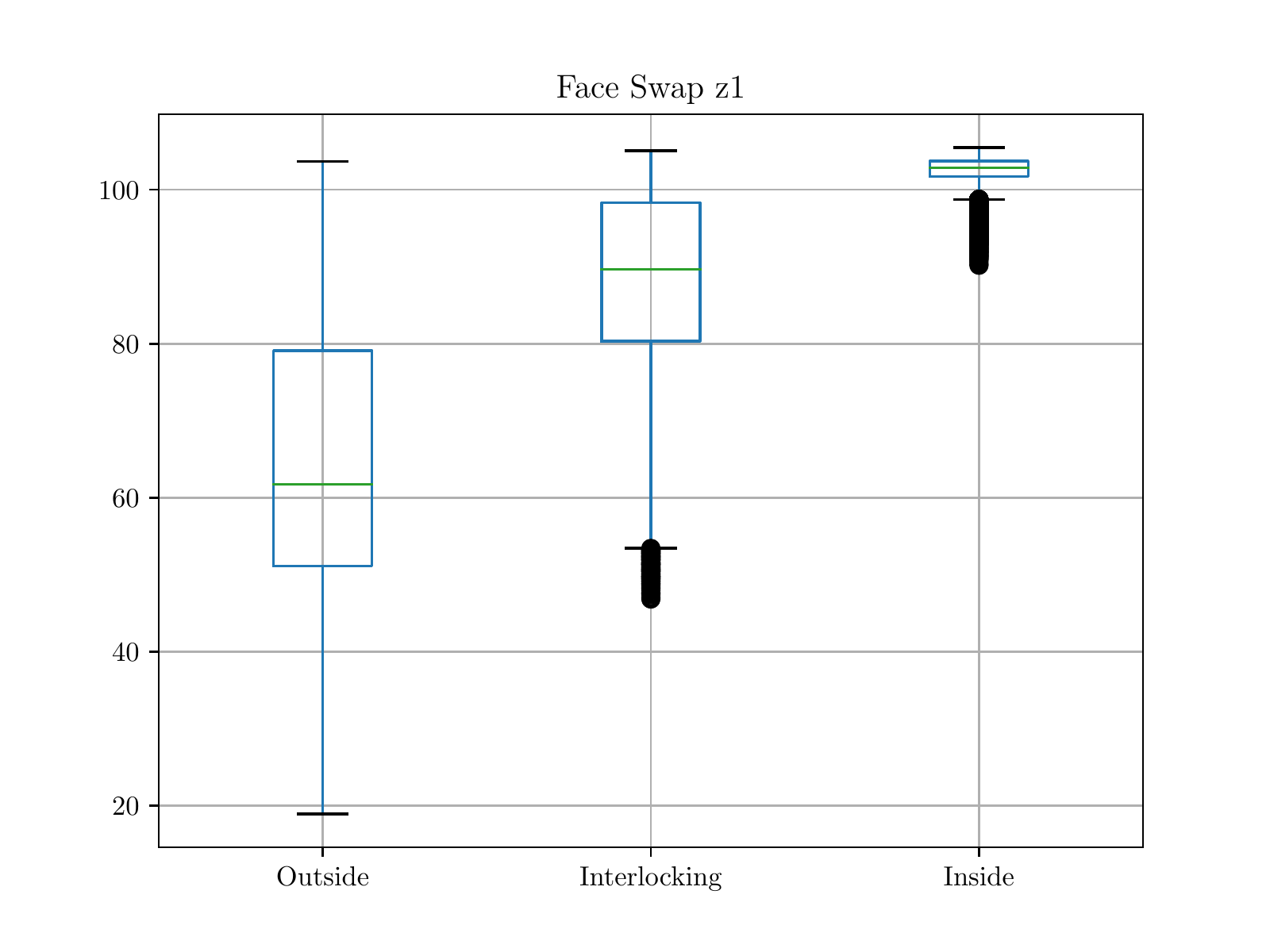}}
\hspace{7pt}
\subfloat{\includegraphics[trim=30 10 40 25, clip, width=0.8in]{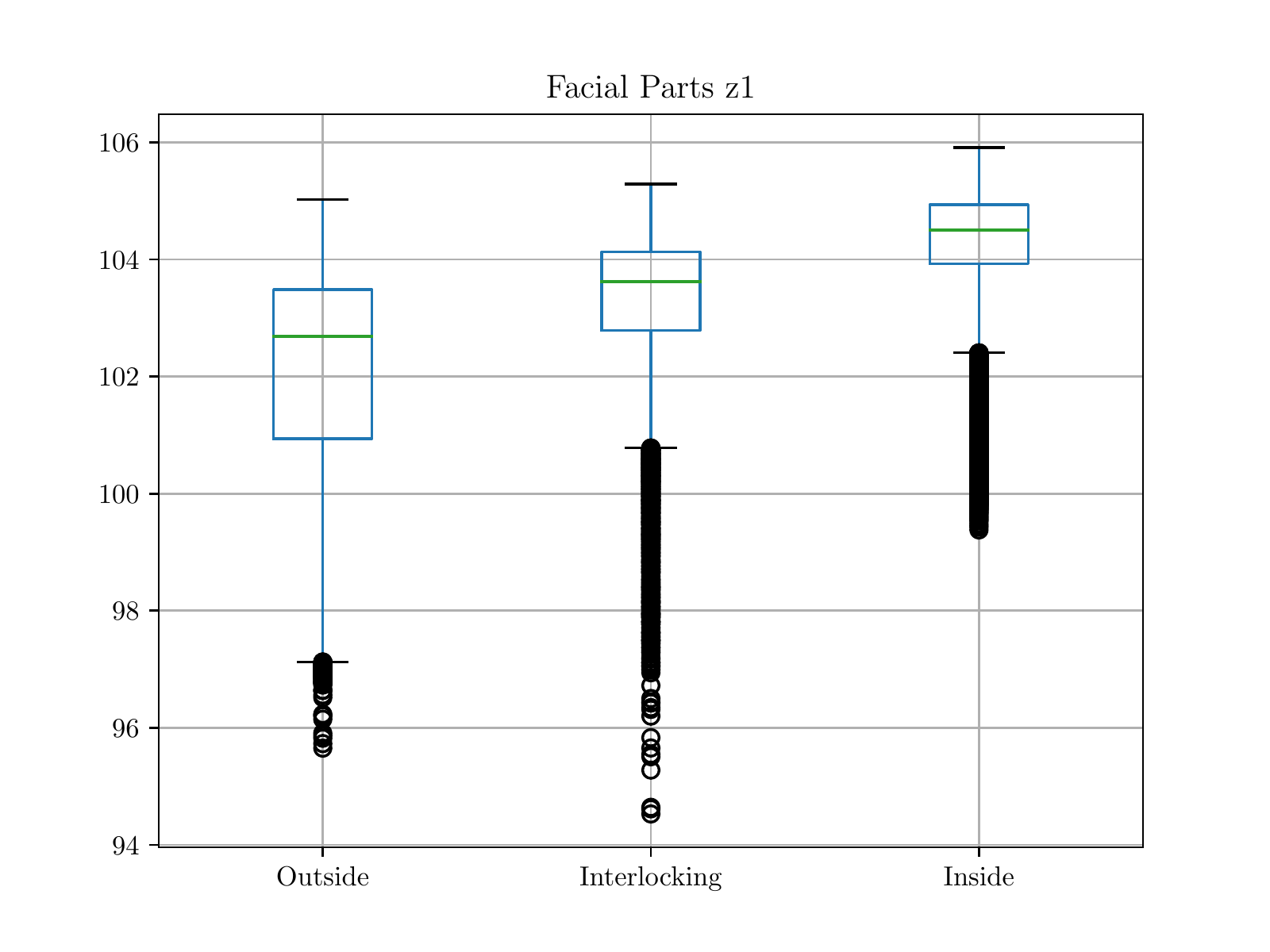}}
\subfloat{\includegraphics[trim=30 10 40 25, clip, width=0.8in]{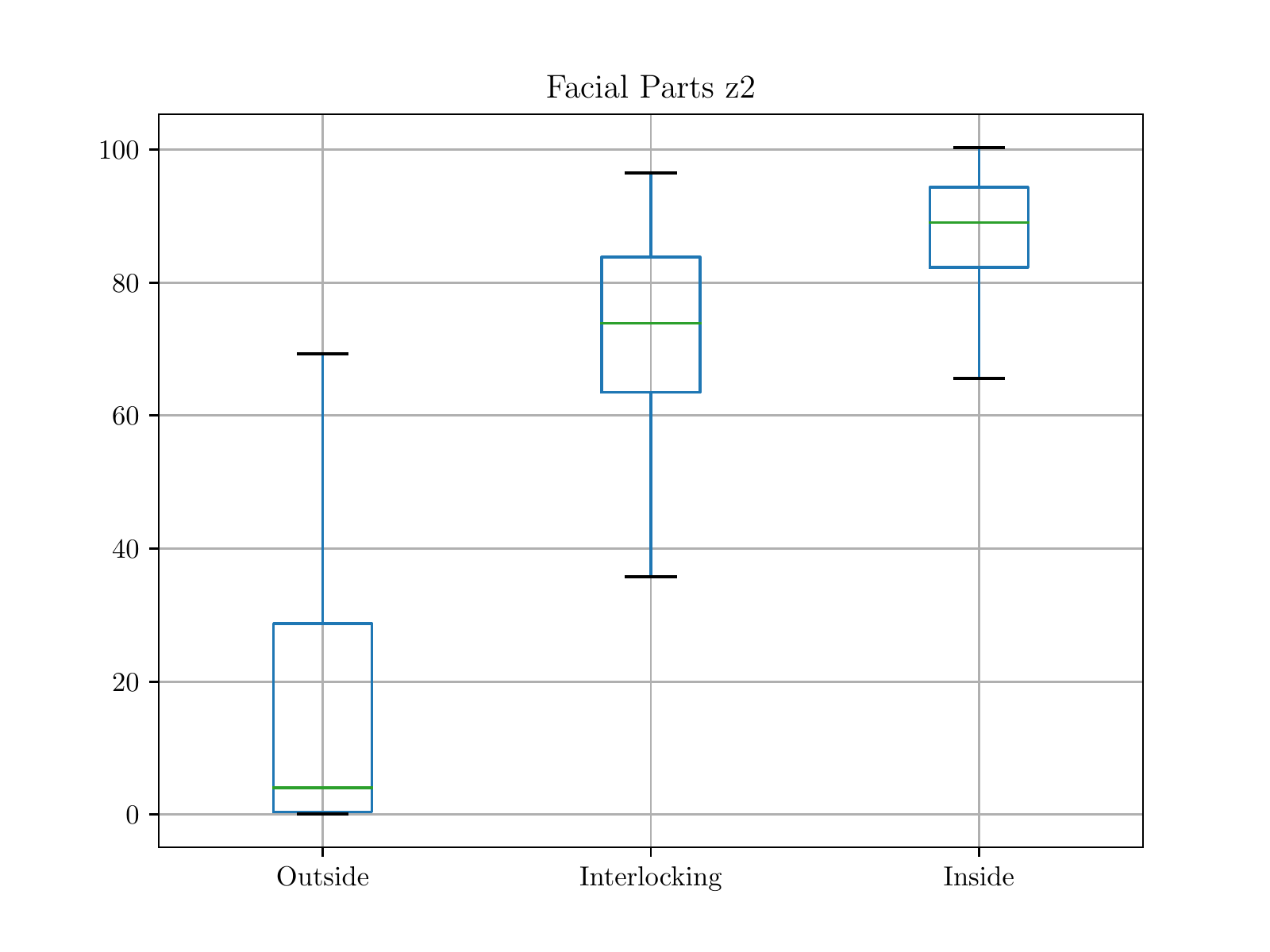}}
\subfloat{\includegraphics[trim=30 10 40 25, clip, width=0.8in]{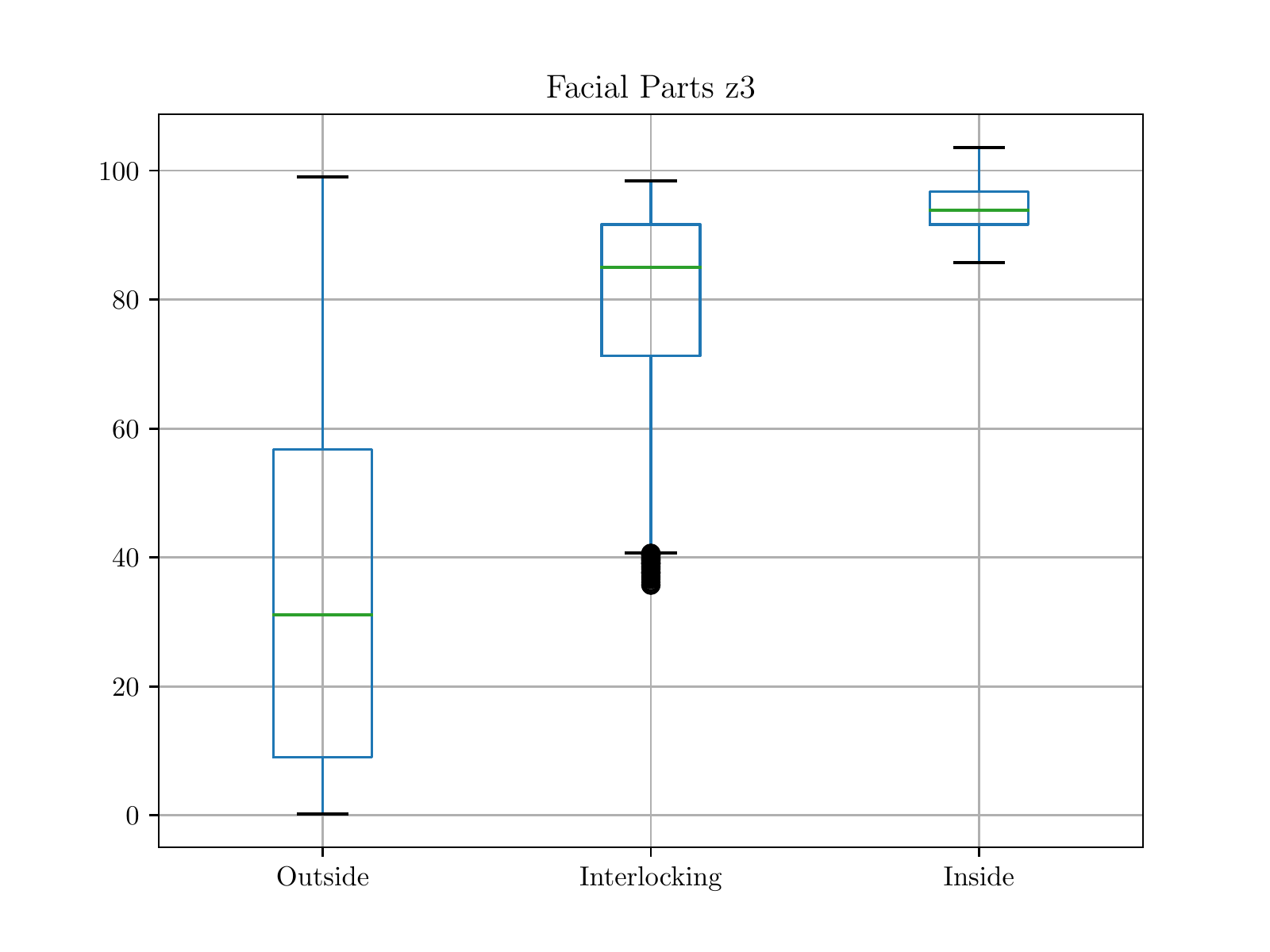}}
\\
\subfloat{\includegraphics[trim=30 10 40 25, clip, width=0.8in]{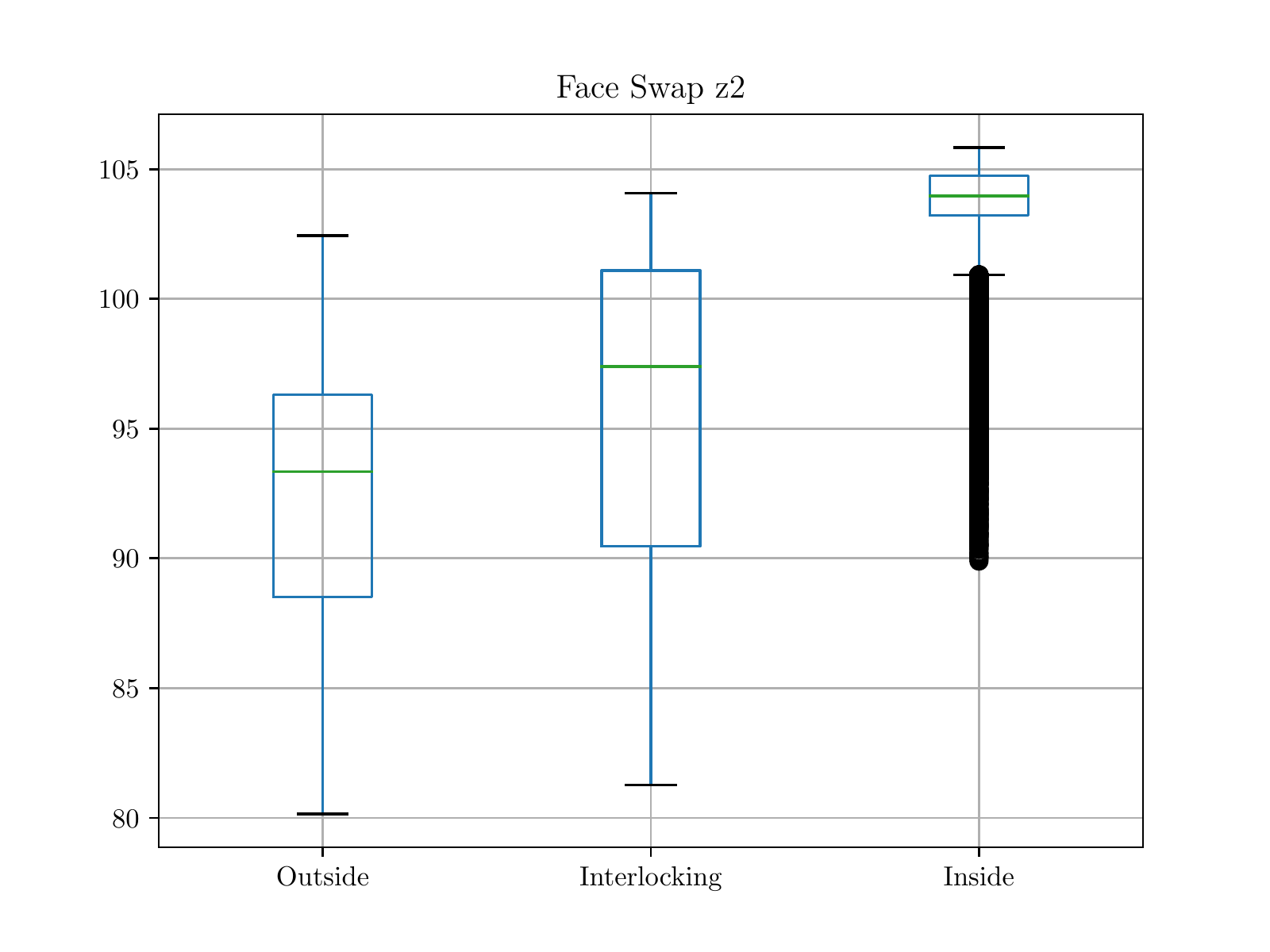}}
\hspace{7pt}
\subfloat{\includegraphics[trim=30 10 40 25, clip, width=0.4in]{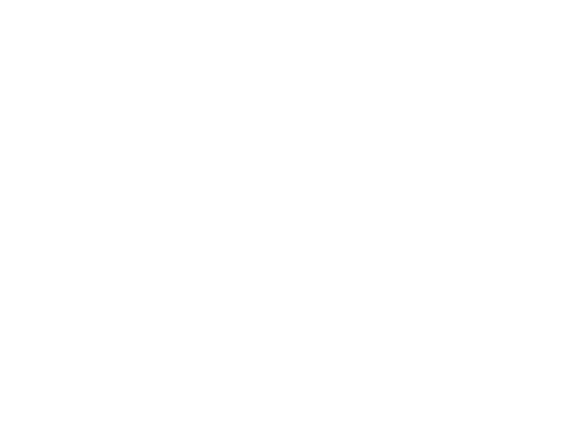}}
\subfloat{\includegraphics[trim=30 10 40 25, clip, width=0.8in]{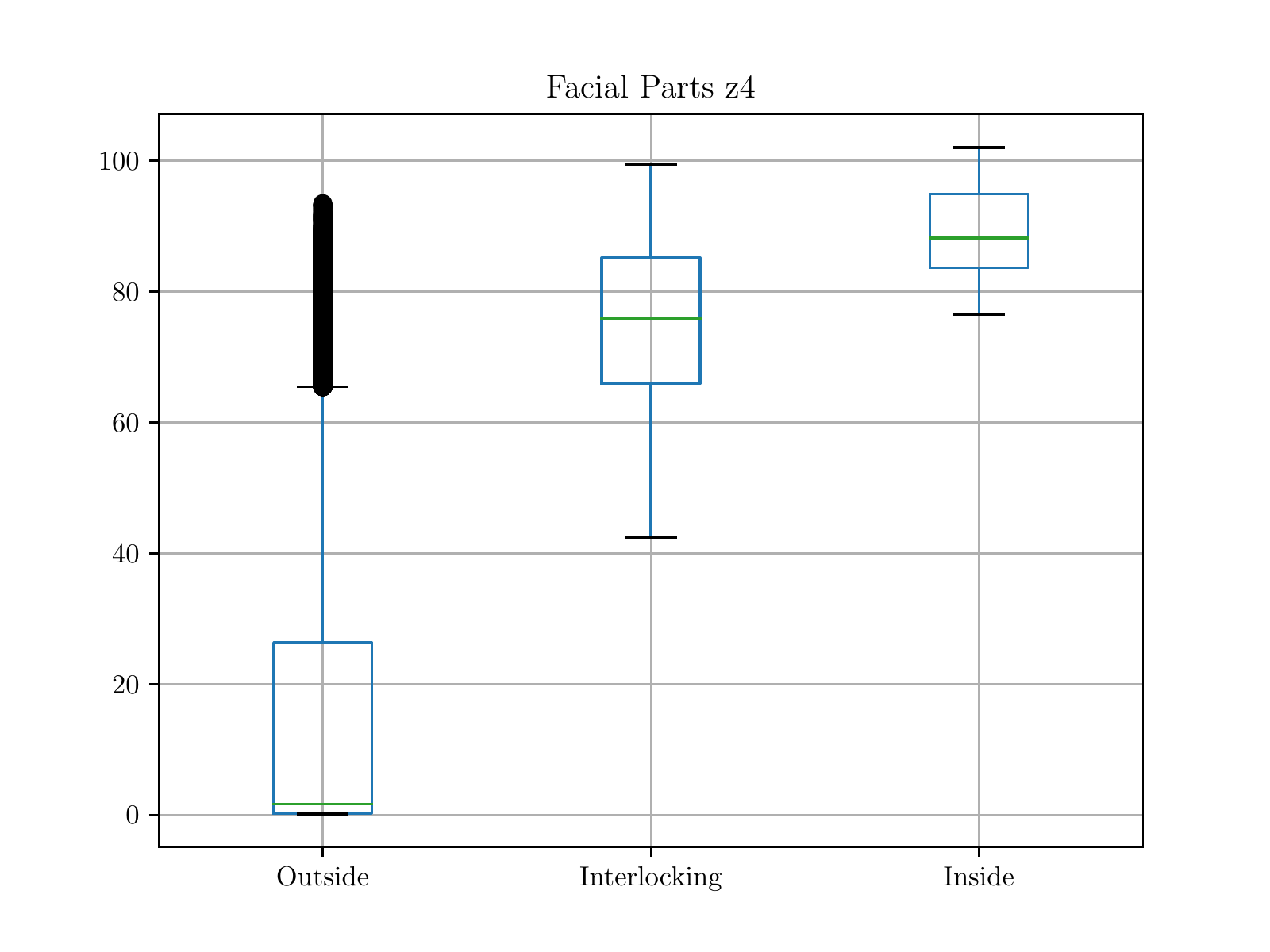}}
\subfloat{\includegraphics[trim=30 10 40 25, clip, width=0.8in]{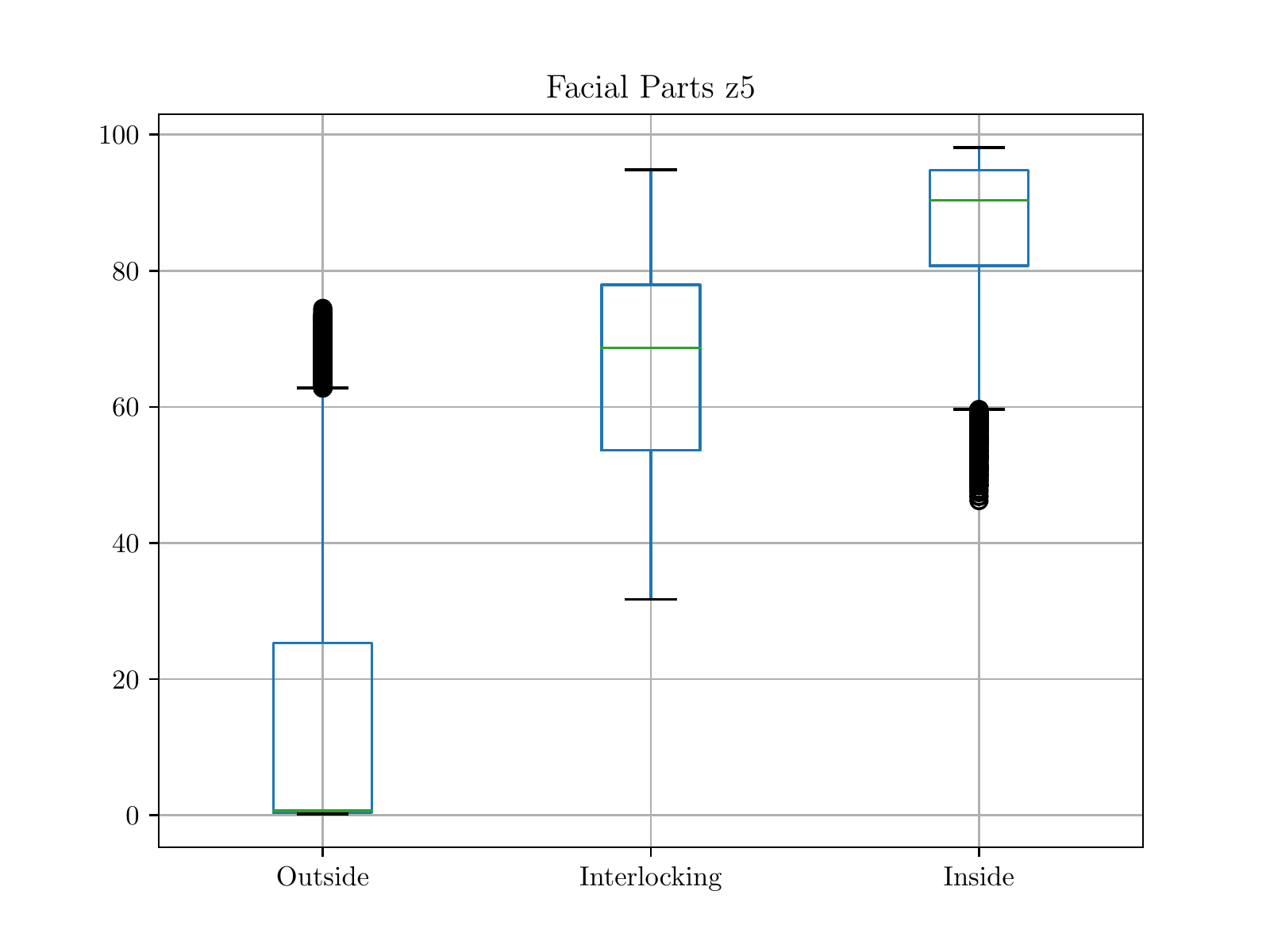}}
\subfloat{\includegraphics[trim=30 10 40 25, clip, width=0.4in]{empty.png}}
\\
	\caption{ Boxplots of MSE in three distinct image regions computed between 50,000 generated images and their edited counterparts by replacing a single $z_i$ only.}
\label{boxplot}
\end{center}
\end{figure}

Figure \ref{boxplot} compares the difference between generated images and their edited counterparts in the three distinct areas; 1) the area inside the influence of replaced $z_i$, 2) the interlocking area that is influenced by $z_i$ and one or more other priors, and 3) the area completely outside the influence of replaced $z_i$. The plots display a decrease in MSE from the inside area, to the interlocking area, and a decrease in MSE from the interlocking area to the outside area.

\section{Conclusion and Future Work}

In this paper we introduce a new scheme in order to make GANs able to learn the distribution of face images as compositions of distributions of smaller parts. We follow the path taken by previous works for making the generated samples by GANs realistic and diverse, and add a new type of structures in the image generating process which leads to disentanglement of concepts based on their positions in the image. This methodology can be applied to various categories of images. The only required modification is to change the way images are divided into parts, which need to be specific to each image category.

Some possible future directions to pursue are combining this framework with advanced detecting and aligning techniques or segmentation techniques for more accurate defining of image parts. Moreover, adding structure in the $z_i$s based on the content that they are responsible to represent may be examined as well.

\bibliographystyle{IEEEtran}

\end{document}